\documentclass[runningheads]{llncs}
\usepackage[breaklinks=true,colorlinks,bookmarks=false]{hyperref}
\usepackage{amssymb}
\usepackage{amsmath}
\usepackage{psfrag}
\usepackage{epsfig}
\usepackage{cite}
\usepackage{graphics}
\usepackage{graphicx}
\usepackage{color}
\usepackage{subfigure}
\usepackage{multirow}
\usepackage{threeparttable}
\usepackage{booktabs}
\usepackage{mmstyles}

\usepackage{bbm}
\usepackage[T1]{fontenc}

\usepackage{booktabs}
\usepackage{bbding}
\usepackage{pifont}

\usepackage{algorithm}
\usepackage{algpseudocode}
\usepackage{graphicx}
\begin{document}
%

\title{Simplify Implant Depth Prediction as Video Grounding: A Texture Perceive Implant Depth Prediction Network}
\titlerunning{TPNet}
%
%



\author{
Xinquan Yang\inst{1,2,3},
Xuguang Li\inst{4},
Xiaoling Luo\inst{1,2,3},
Leilei Zeng\inst{1,2,3},
Yudi Zhang\inst{1,2,3},
Linlin Shen\inst{1,2,3}$^{\dagger}$,
Yongqiang Deng\inst{4}$^{\dagger}$,
}

\institute{College of Computer Science and Software Engineering, Shenzhen University, Shenzhen, China \\ \and AI Research Center for Medical Image Analysis and Diagnosis, Shenzhen University, Shenzhen, China \and National Engineering Laboratory for Big Data System Computing Technology, Shenzhen University, China \and Department of Stomatology, Shenzhen University General Hospital, Shenzhen, China \\
\email{xinquanyang99@gmail.com, llshen@szu.edu.cn, qiangyongdeng@sina.com}
}
{\let\thefootnote\relax\footnotetext{$^{\dagger}$ Corresponding Author}}

%
\maketitle              
\begin{abstract}
Surgical guide plate is an important tool for the dental implant surgery. However, the design process heavily relies on the dentist to manually simulate the implant angle and depth. When deep neural networks have been applied to assist the dentist quickly locates the implant position, most of them are not able to determine the implant depth. Inspired by the video grounding task which localizes the starting and ending time of the target video segment, in this paper, we simplify the implant depth prediction as video grounding and develop a Texture Perceive Implant Depth Prediction Network (TPNet), which enables us to directly output the implant depth without complex measurements of oral bone. TPNet consists of an implant region detector (IRD) and an implant depth prediction network (IDPNet). IRD is an object detector designed to crop the candidate implant volume from the CBCT, which greatly saves the computation resource. IDPNet takes the cropped CBCT data to predict the implant depth. A Texture Perceive Loss (TPL) is devised to enable the encoder of IDPNet to perceive the texture variation among slices. Extensive experiments on a large dental implant dataset demonstrated that the proposed TPNet achieves superior performance than the existing methods.
\end{abstract}

\section{Introduction}
Tooth loss is a common problem among middle-aged and elderly people, and artificial dental implantation is one of the most appropriate treatment methods. 
\begin{figure}
\centering
\includegraphics[width=0.95\linewidth]{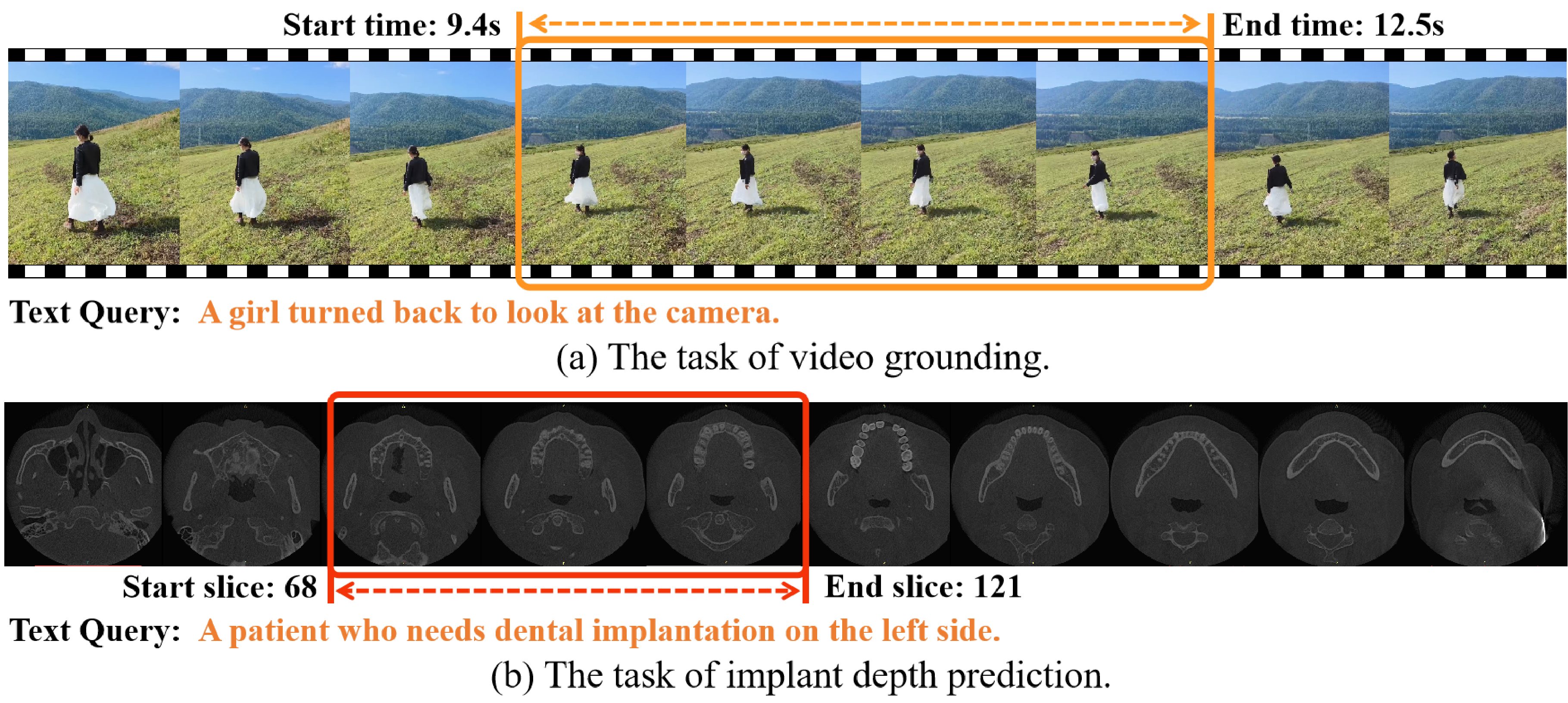}
\caption{Comparison of the video grounding task and implant depth prediction task. 
} 
\label{fig_vg_idp}
\end{figure}
In clinical, to ensure implant accuracy and accelerate the implantation process, dentists usually use the surgical guide plate during surgery. However, the design of the surgical guide plate require to manually simulate the implant position (e.g., implantation angle and depth) by loading the Cone-beam computed tomography (CBCT) data into the design software, which is labour-intensive and time-consuming. With the development of deep learning, using artificial intelligence methods to speed up such process is promising. 

Recently, a number of literature works have been proposed to assist the dentist quickly locating the implant position. ImplantFormer~\cite{yang2024implantformer} proposed to predict the implant position using the 2D axial view of tooth crown images and projects the prediction results back to the tooth root by the space transform algorithm. Following this paradigm, a series of improved works, TSIRP~\cite{yang2024two}, TCEIP~\cite{yang2023tceip}, and TCSloT~\cite{yang2023tcslot} are proposed to improve the accuracy of implant position prediction. Although these methods demonstrate excellent performance, they are semi-automated as the dentist are required to manually set the implant depth, which is inefficient for the clinic application. To solve this problem, some researchers try to detect the alveolar bone and mandibular canal based on the sagittal view of CBCT to determine the height and width of the alveolar bone, which predicts a approximate implant depth~\cite{widiasri2022dental}. Kurt et al.~\cite{kurt2021deep} utilised multiple pre-trained convolutional networks to segment the teeth and jaws to locate the missing tooth and determine the implant depth by measuring oral tissues (e.g., mandibular canal, maxillary sinus, and jaw bone edge). However, these methods are too complicated for the clinic application and can not provide a precise implant depth.

Video grounding is an important yet challenging task in computer vision, which requires the machine to watch a video and localize the starting and ending time of the target video segment that corresponds to the given query~\cite{zeng2020dense}. In this paper, we found that the task of implant depth prediction is similar to video grounding, if we consider the 3D CBCT data as a video and the beginning and ending slice of implant as the starting and ending time of the target video segment, as shown in Fig.~\ref{fig_vg_idp}. 
\begin{figure}
\centering
\includegraphics[width=0.95\linewidth]{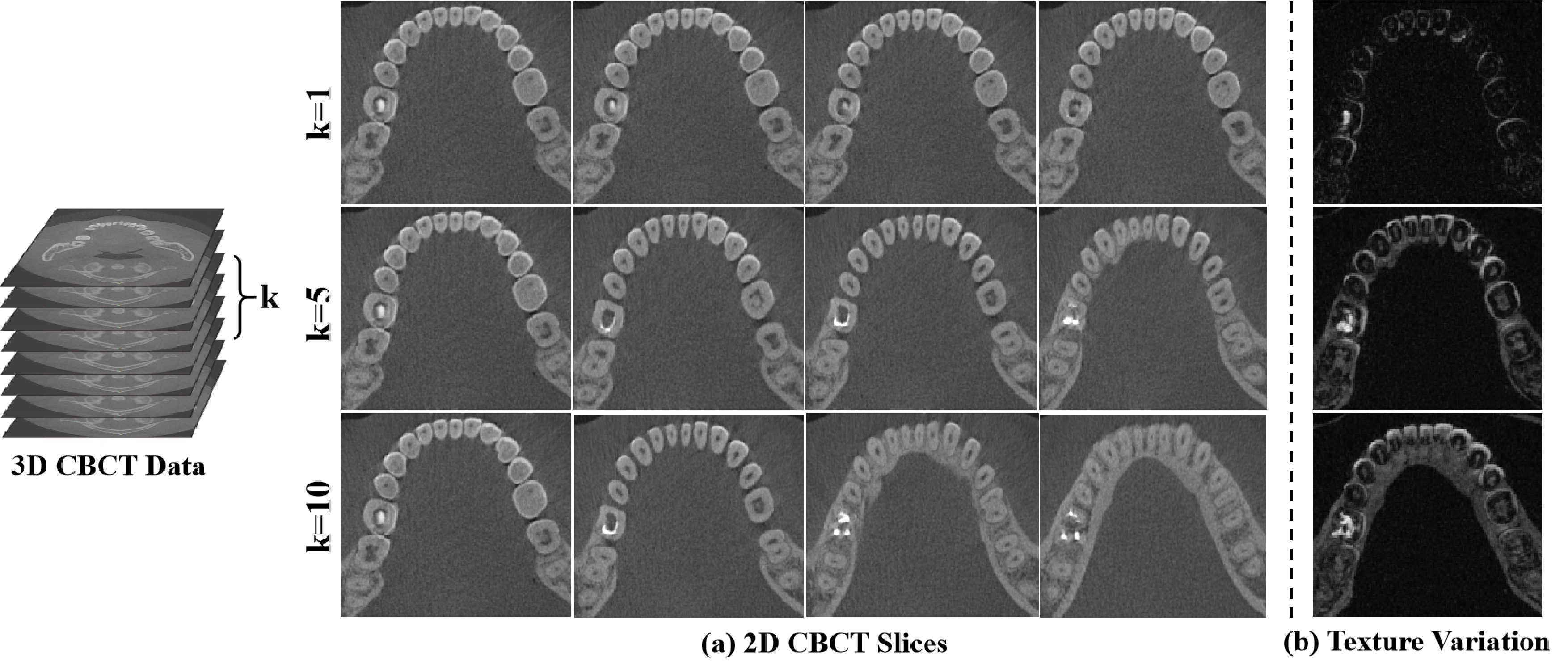}
\caption{(a) Comparison of 2D slices with different sampling intervals, $k$ represents the sampling interval. (b) Texture variation computed from the 2D slices.
} 
\label{fig_gap}
\end{figure}
By this means, the implant depth can be directly determined during inference, without requiring additional measurements of oral tissues.

Motivated by the above observation, in this paper, we develop a Texture Perceive Implant Depth Prediction Network (TPNet), which consists of an implant region detector (IRD) and an implant depth prediction network (IDPNet). IRD is an object detector designed to locate the implant region. We crop a sub-volume from the CBCT data according to the detection result of IRD. By this means, the irrelevant information of CBCT for implantation will be removed and the input data size can be substantially reduced. Then, the sub-volume is taken as the input of IDPNet. IDPNet is devised to regress the precise implant depth, which is a single encoder-decoder regression network. As the determination of implant depth relies on the texture of neighboring teeth, a Texture Perceive Loss (TPL) is proposed to enable the encoder to perceive the texture variation among slices, which greatly helps the IDPNet predicts more accurate implant depth.

Main contributions of this paper can be summarized as follows:1) To the best of our knowledge, we are the first one to model the task of implant depth prediction as video grounding, which enables us to directly predict the implant depth, without requiring additional computation. 2) An implant region detector (IRD) is introduced to remove the irrelevant information of CBCT, which sharply reduces the input data size and save computational costs. 3) A Texture Perceive Loss (TPL) is devised to enable the encoder to capture more fine-grained features by perceiving the texture variation among slices. 4) Extensive experiments on a large dental implant dataset demonstrated the proposed TPNet achieves superior performance than the existing methods.

\begin{figure}
\centering
\includegraphics[width=0.95\linewidth]{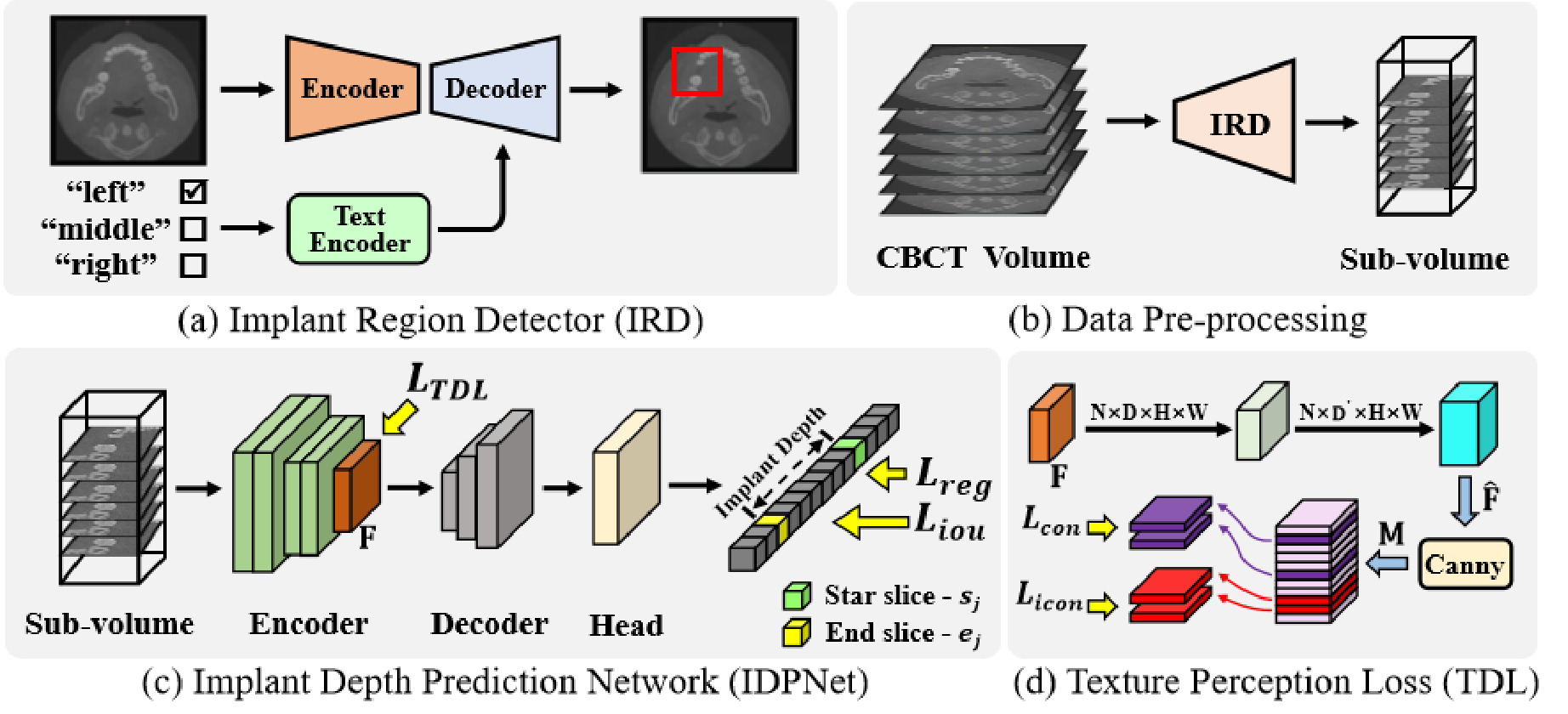}
\caption{The architecture of the proposed texture perceive implant depth prediction framework.} \label{fig_network}
\end{figure}

\section{Method}
Given a patient's CBCT data, TPNet aims to predict a precise implant depth, i.e., the index of start slice and end slice. An overview of TPNet is presented in Fig.~\ref{fig_network}. It mainly consists of two parts: i) Implant Region Detector (IRD), ii) Implant Depth Prediction Network (IDPNet). IRD first locates the implant region to crop a sub-volume from the CBCT data, and the IDPNet takes the sub-volume as input to predict the implant depth. Next, we will introduce them in detail.

\subsection{Implant Region Detector}
The CBCT data contains complete information about the maxillary and mandibular bones, in which the maxillary and mandibular sinuses are irrelevant for predicting implant depth. Therefore, it is computationally intensive to train IDPNet using the whole CBCT data. Using an IRD to detect the implant region and crop a sub-volume according to the detection result can significantly reduce the CBCT size. Inspired by the previous methods, we introduce a text guided implant position prediction network - TCEIP~\cite{yang2023tceip} as IRD. TCEIP integrates the direction embedding form CLIP~\cite{radford2021learning} to guide the prediction model to locate the implant position, thus perform well in the patient who have multiple missing teeth. Consider clinical practicality, in this paper, we design a lightweight TCEIP as IRD, in which the knowledge alignment module and cross-modal attention module are discarded. 

The architecture of IRD is shown in Fig.~\ref{fig_network}(a), which consists of an encoder, a decoder and a text encoder of CLIP. Firstly, ResNet-50~\cite{he2016deep} is used as the encoder for feature extraction, and three deconvolution layers are adopted as the decoder to recover high-resolution features. Then, we extract the conditional text embedding from CLIP by inputting an additional text, e.g., 'left', 'middle', or 'right' into the CLIP text encoder. In the end, the conditional text embedding is concatenated with the last feature map of decoder to generate a gaussian heatmap for implant position regression. We follow TCEIP to use the focal loss~\cite{lin2017focal} and L1 loss for supervision. After obtaining the implant position, we generate a $256\times256$ box centered on the implant position as the implant region, which ensures that the texture of neighboring teeth is included. We then crop a $352\times256\times256$ sub-volume from the original CBCT along the axial view, and the sub-volume will be taken as the input data of IDPNet.

\subsection{Implant Depth Prediction Network}
The architecture of IDPNet is shown in Fig.~\ref{fig_network}(c). It mainly consists of an encoder, a decoder and a regression head. Firstly, the encoder extracts features from the sub-volume and the middle feature map $\mF\in \mathbb{R}^{N\times C\times D\times H\times W}$ will be extracted. We use the proposed Texture Perception Loss (TPL) to supervise $\mF$ that enables the encoder can capture more fine-grained features by perceiving the texture variation among slices. Then, the decoder recovers the encoder features to high-resolution and the regression head predicts the implant depth. Next, we will introduce them in detail.

\subsubsection{Encoder and Decoder.}
We employ the widely used resblock to construct the encoder of IDPNet. Specifically, the encoder consists of two 3D resblocks~\cite{tran2018closer} and two 2D resblocks~\cite{he2016deep}. The architecture of encoder and decoder are shown in the Fig.~\ref{fig_network}(c). The 3D resblock first takes the sub-volume as input and learns context information among slices. Then, these temporal features are fed into the 2D resblock to learn the texture feature in different slices. The output of encoder is a feature map $\mF\in \mathbb{R}^{N\times C\times D\times H\times W}$. Considering that the regression of implant depth heavily relies on clearly neighboring tooth texture, which requires high-resolution feature representations. Hence, we adopt three deconvolution layers as decoder to consecutively upsamples the feature map.

\subsubsection{Texture Perceive Loss.}
Clinically, dentists determine the implant depth according to the texture of neighboring teeth, e.g., the bottom of the implant does not exceed the root of neighboring teeth. Therefore, IDPNet should possess the ability to perceive the texture variation among slices. In Fig.~\ref{fig_gap}, we visualize 2D slices sampled with different sampling and compute the texture variation among these slices by standard deviation. We can observe from the figure that the larger the sampling interval, the more obvious such texture variations. This observation indicates that the neighboring 2D slices have a similar feature, while the distant slices have a big difference in features. Drawing inspiration from this observation, in this paper, we propose a Texture Perceive Loss (TPL), which assists the encoder learns more robust features. 

The details of TPL is given in the Fig.~\ref{fig_network}(d). Specifically, we first reduce the channel $C$ of $\mF$ to 1 and reshape the channel $D$ to $D^{'}$, to restore the information of channel $D$. By this means, the pre-processed $\hat{\mF}\in \mathbb{R}^{N\times D^{'}\times H\times W}$ is obtained. Then, we apply the Canny operator for $\hat \mF$ to extract textures along the channel $D^{'}$. After obtaining a series of texture matrix $\mM\in \mathbb{R}^{N\times D^{'}\times H\times W}$, we perform the consistency loss $\mathcal{L}_{con}$ for the neighboring matrix to close these features, and the inconsistency loss $\mathcal{L}_{icon}$ for the distant matrix to distinguish these features. $\mathcal{L}_{TPL}$ is the summation of $\mathcal{L}_{con}$ and $\mathcal{L}_{icon}$, and we implement $\mathcal{L}_{con}$ and $\mathcal{L}_{icon}$ by L2 loss. In our implementation, we set the sampling interval $k$ of the distant matrix as 10. 


\subsubsection{Regression Head.}
The regression head is designed to predict the implant depth, which is implemented by two convolutions followed by the activation function, i.e., ReLU. We use the L1 loss to optimize the regression head:
\begin{equation}
    \mathcal{L}_{reg} = \sum_{j=1}^{N_p} |y_j - \hat y_j|, 
\end{equation}
where $j$ is the patient index in a mini-batch and $N_p$ is the total number of patient. $y_j=(s_j,e_j)$ and $\hat y_j=(\hat s_j,\hat e_j)$ is the predicted and ground-truth index of start and end implant slice, respectively. 

As discussed in previous sections, we model implant depth prediction as the task of video grounding. Therefore, we follow the video grounding to introduce the temporal iou loss~\cite{zhang2023text} to supervise the regression head:
\begin{equation}
    \mathcal{L}_{tiou} = 1-\frac{\hat y_j\cap y_j}{\hat y_j\cup y_j}. 
\end{equation}

The rationale of $\mathcal{L}_{tiou}$ is to maximize the overlapping between the predicted slice index and its ground truth. The overall training loss of IDPNet is:
\begin{equation}
\mathcal{L}_{total}=\mathcal{L}_{reg}+\mathcal{L}_{tiou}+\mathcal{L}_{TPL}
\end{equation}

\begin{table}
\caption{The ablation experiments of each components in IRD.}\label{table_IRD}
\centering
\begin{tabular}{c|c|c|c|c} 
\toprule
Network                  & Knowledge Alignment         & Cross-modal Attention & $AP_{75}$$\uparrow$  & FLOPs(G$)\downarrow$ \\ \hline
\multirow{4}{*}{IRD}     & \Checkmark  & \Checkmark       & \textbf{18.4}    & 67.48\\
                         & \Checkmark  & \ding{55}        & 17.1             & 56.88\\
                         & \ding{55}   & \Checkmark       & 16.8             & 66.81\\
                         & \ding{55}   & \ding{55}        & 16.2             & \textbf{56.21}\\
\bottomrule
\end{tabular}
\end{table}

\section{Experiment}\label{sec6}
\subsection{Dataset and Implementation Details}
We evaluate the proposed TPNet on a large dental implant dataset, which were collects from the Shenzhen University General Hospital (SUGH). The dataset contains 400 patients, in which 80\% data were selected as the training set and the remaining 20\% as the testing set. All the CBCT data were captured using the KaVo 3D eXami machine, manufactured by Imagine Sciences International LLC. The original CBCT size is $432\times 776\times 776$. For the traing of IRD, we follow TCEIP to use the 2D slice of CBCT and resize them to $512\times 512$ for training and inference. After the data pre-processing of IRD, the size of CBCT data for each patient is reduced to $352\times 256\times 256$. 

For the training of IRD, we use a batch size of 8, Adam optimizer and a learning rate of 0.001. Total training epochs is 80 and the learning rate is divided by 10 when epoch $=\{40, 60\}$. Three data augmentation methods, i.e. random crop, random scale and random flip are employed. For the training of IDPNet, we use a batch size of 1, SGD optimizer and a learning rate of 0.001.
As the asymmetric structure of the upper and lower jaws, only the horizontal flip is applied for data augmentation. IDPNet is trained for 40 epochs and the learning rate is divided by 10 at 20th and 30th epochs, respectively. All the models are trained and tested on the platform of NVIDIA A100 GPU.




\begin{table}[]
\centering
\caption{Performance comparison of different loss function.}\label{ablation_loss}
\scalebox{0.9}{
\begin{tabular}{ccc|ccc}
\toprule  
\multirow{2}{*}{$\mathcal{L}_{reg}$} & \multirow{2}{*}{$\mathcal{L}_{tiou}$} & \multirow{2}{*}{$\mathcal{L}_{TPL}$} & \multicolumn{3}{|c}{Acc(R@1, IoU=m)}  \\
                         &  &  & m=0.6      & m=0.7      & m=0.8    \\ \hline
\Checkmark             & \ding{55}  &  \ding{55}                   &  28.8        & 23.7       &  15.3         \\
 \Checkmark            & \Checkmark & \ding{55}          & \textbf{35.6}    & \textbf{28.8}         &  16.9           \\
 \Checkmark            & \Checkmark & \Checkmark & 33.9    & 25.4         & \textbf{20.3}          \\ 
\bottomrule
\end{tabular}
}
\end{table}

\subsection{Performance Analysis}
In the task of implant depth prediction, the kernel of the implant should not invade the mandibular nerve canal and should maintain a minimum safety distance of 1.5mm. Therefore, as long as the center point of the implant root conforms to this rule, it is a good prediction. In this paper, we consider the timeline of the video as the sagittal axis of CBCT, so we can directly use IOU to measure the accuracy of implant depth prediction while ensuring that the implant roots meet the standards(>1.5mm). We follow previous work~\cite{gao2017tall} to adopt Acc(R@1, IoU=m) as the performance evaluation metric, which represents the percentage accuracy of top-1 predicted moments whose IoU with the ground-truth moment is larger than m. We set the IoU threshold values m=\{0.6, 0.7, 0.8\}. 

\begin{figure}
\centering
\includegraphics[width=0.9\linewidth]{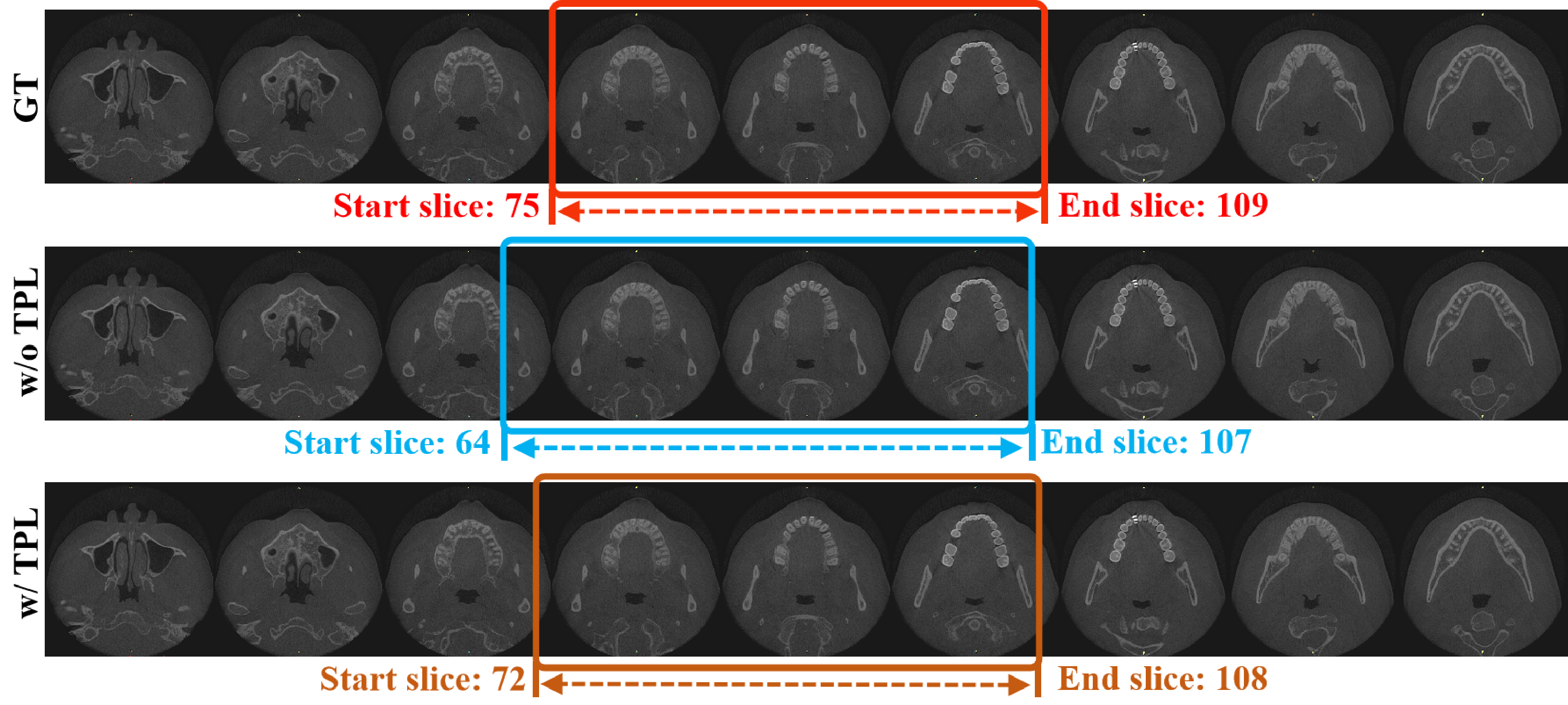}
\caption{Detection results of TPNet trained with or without TPL loss.} \label{fig_compare}
\end{figure}

\begin{table}[]
\caption{Performance comparison to the video grounding methods.}\label{table_compare}
\centering
\scalebox{0.9}{
\begin{tabular}{c|c|ccc}
\toprule
\multirow{2}{*}{Method} & \multirow{2}{*}{Visual Feature} & \multicolumn{3}{c}{Acc(R@1, IoU=m)}\\ 
                        &     & m=0.6      & m=0.7      & m=0.8     \\ \hline
TSP-PRL~\cite{wu2020tree}           &  C3D   &  33.1     &  \textbf{26.8}       & 18.6          \\
MAN~\cite{zhang2019man}             &  I3D   &  32.6     &  23.1        &  15.8         \\
VSLNet~\cite{zhang2020span}         &  I3D   &  31.2      &  23.5      &  17.1         \\
DRN~\cite{zeng2020dense}            &  I3D   &  \textbf{34.5}      &  21.7      &  16.3         \\
TPNet(ours)             &  -        & 33.9    & 25.4         & \textbf{20.3}           \\
\bottomrule
\end{tabular}
}
\end{table}

\subsubsection{Ablation Studies of IRD.} As the IRD is used as pre-processing method to crop CBCT data  for IDPNet, an approximate planting area is sufficient but requires quick inference speed. To evaluate the effectiveness of the proposed IRD, we conduct ablation experiments to investigate the effect of removing components in IRD, results are given in Table~\ref{table_IRD}. $AP_{75}$ and FLOPs are used as evaluation metrics. We can observe from the table that removing both modules will result in a 2.2\% performance decrease, but FLOPs is decreased by 11.27. This results meet with the requirement and demonstrate that the proposed IRD is effective and lightweight for clinical practice. 

\subsubsection{Ablation Studies of Loss Function.} To demonstrate the effectiveness of the proposed loss function, we conduct ablation experiments to investigate the effect of each loss function in Table~\ref{ablation_loss}. We can observe from the second row of the table that using temporal iou loss alone will lead to regression failure. When combining both regression loss and temporal iou loss, the accuracy (m=0.8) improves by 1.6\%. When the TPL loss is introduced, the improvement reaches to 5.0\%. Although the accuracy of smaller IoU thresholds has decreased, high IoU threshold is required in clinical practice. This results demonstrate the effectiveness of TPL loss, which enables the encoder to perceive the texture variation among slices.  

\subsubsection{Visual Comparison.} To further validate the effectiveness of the proposed TPL loss, in Fig.~\ref{fig_compare}, we visualize the prediction result of TPNet with or without training by the TPL loss. From the figure we can observe that the introduction of TPL predict more precise start and end slices of the implant, due to the perception capability of texture variation. 

\subsubsection{Comparison to the Video Grounding Methods.} As previously discussed, we model the task of implant depth prediction as video grounding. To demonstrate the superior performance of the proposed method, we compare TPNet with other state-of-the-art video grounding methods in Table~\ref{table_compare}. Specifically, we choose different visual feature based methods, e.g., the C3D-based method, TSP-PRL, and the I3D-based methods, MAN, VSLNet and DRN. From the table we can observe that, the C3D-based method perform better than the I3D-based networks in high iou threshold (e.g., TSP-PRL achieved 18.6\% Acc, which is 1.5\% higher than the best-performing I3D-based network - VSLNet). The proposed TPNet achieves the best accuracy of 20.3\%, among all benchmarks. The experimental results proved the effectiveness of our method.

\section{Conclusions}
In this paper, we simplify the task of implant depth prediction as video grounding, and develop a texture perceive implant depth prediction network (TPNet). TPNet consists of an implant region detector (IRD) and an implant depth prediction network (IDPNet). IRD is an object detector designed to reduce the size of CBCT by cropping a probable implant region from CBCT data. IDPNet is devised to regress the precise implant depth. A texture consistency (TC) loss is designed to enable the image encoder to capture more fine-grained features. Extensive experiments on a large dental implant dataset demonstrated that the proposed TPNet achieves superior performance than the existing methods.

\bibliographystyle{splncs04}
\bibliography{ref}
\end{document}